\documentclass{article}

\usepackage{arxiv}

\usepackage{subcaption}
\usepackage[utf8]{inputenc} % allow utf-8 input
\usepackage[T1]{fontenc}    % use 8-bit T1 fonts
\usepackage{hyperref}       % hyperlinks
\usepackage{url}            % simple URL typesetting
\usepackage{booktabs}       % professional-quality tables
\usepackage{amsfonts}       % blackboard math symbols
\usepackage{nicefrac}       % compact symbols for 1/2, etc.
\usepackage{microtype}      % microtypography
\usepackage{lipsum}
\usepackage{graphicx}
\graphicspath{ {./images/} }

\title{A comparative approach to assessing linguistic creativity of Large Language Models and Humans}

\author{
 Anca Dinu \\
  University of Bucharest\\
  Bucharest, Romania  \\
  \texttt{anca.dinu@lls.unibuc.ro} \\
   \And
 Andra-Maria Florescu \\
  Interdisciplinary School of Doctoral Studies\\
  University of Bucharest\\
  Bucharest, Romania \\
  \texttt{andra-maria.florescu@s.unibuc.ro} \\
  \And
 Alina Resceanu \\
  University of Craiova\\
  Craiova, Romania \\
  \texttt{alina.resceanu@edu.ucv.ro} \\
}

\begin{document}
\maketitle
\begin{abstract}
The following paper introduces a general linguistic creativity test for humans and Large Language Models (LLMs). The test consists of various tasks aimed at assessing their ability to generate new original words and phrases based on word formation processes (derivation and compounding) and on metaphorical language use. We administered the test to 24 humans and to an equal number of LLMs, and we automatically evaluated their answers using OCSAI tool for three criteria: \textit{Originality}, \textit{Elaboration}, and \textit{Flexibility}. The results show that LLMs not only outperformed humans in all the assessed criteria, but did better in six out of the eight test tasks. We then computed the uniqueness of the individual answers, which showed some minor differences between humans and LLMs. Finally, we performed a short manual analysis of the dataset, which revealed that humans are more inclined towards E(extending)-creativity, while LLMs favor F(ixed)-creativity. 
\end{abstract}

% keywords can be removed

\section{Introduction and Theoretical Background}
\label{intro}

Creativity represents a human’s innate ability to create, based on preexisting knowledge and experience, something innovative and viable \cite{carayannis2013encyclopedia}. Guilford \cite{Guilford67} stated that there are two main types of thinking in the creative process: divergent thinking, which refers to the plethora of ideas that occur when faced with a creative task, and convergent thinking, which limits these ideas to only the most suitable ones. Both are needed for a successful creative process and are implicitly contained in the notion of creativity. Not long ago, creativity was considered a human-specific trait. Computational creativity is a relatively new domain that originated in 1960 and exploded in the last five years, with the rise of AI and Large Language Models. The various types of creativity, such as ideational creativity, problem solving creativity, writing creativity, linguistic creativity, figural (imagistic) creativity, musical creativity, etc. are now all exhibited by machines, not by humans only. The focus of this article is the linguistic creativity of humans and machines alike.

Linguistic creativity is, most generally, the ability of an individual to use natural language in new and unusual ways. It is deeply embedded in our everyday communication \cite{jones2015routledge}, in the way we use and manipulate language, and in the continuous flow of the media we consume. Its omnipresence and rapid growth have made linguistic creativity a subject of extensive study in numerous disciplines, from psychology, neuroscience, computer science \cite{Boden_2009}, to linguistics \cite{Vasileanu2024KeepCA}, history, criticism, and even management theory
\cite{carayannis2013encyclopedia, Kaufman_Sternberg_2010, Gaut2003}. 

This multidimensional nature highlights the fact that what constitutes 'linguistic creativity' varies fundamentally based on the different perspectives of analysis. Sampson \cite{sampson2016two} distinguishes between two types of linguistic creativity: F-creativity (Fixed) and E-creativity (Extending), which do not form a discrete space, but a rather continuous one. An instance of F-creativity definition is Chomsky's: “Language is a process of free creation; its laws and principles are fixed, but the manner in which the principles of generation are used is free and infinitely varied” \cite{chomsky2003reasons}. This kind of creativity refers to the rule-based syntactic productivity, morphological productivity, and semantic productivity (like snow-clones).  Bergs' definition of creativity focuses on E-creativity: "It is this truly creative in the sense that we create something original, unexpected, novel, unheard, unproduced" \cite{Bergs2019173183}. It refers to the language uses that break the morpho-grammatical and semantic rules, like syntactic or semantic mismatches, such as in the use of metaphors, metonymy, or other literary devices. Both F-creativity and E-creativity are relevant for this work. 

Recent research on human linguistic creativity pointed out that young people are the main drivers of originality and change in language use, generating new words for new concepts (such as \textit{ mansplaining}), slang (\textit{brain rot}), and complex formations like nonce words, defined in \cite{Bauer_1983} as "new complex words created by a speaker/writer on the spur of the moment to cover some immediate need" (such as \textit{supercalifragilisticexpialidocious}), with remarkable ease and frequency. They treat language as a playground, bypassing swear words (\textit{duck you}) and slurs filters (\textit{restarted} instead of \textit{retarded}), prioritizing expressivity, humor, and in-group identity over rigid morphological rules \cite{Tagliamonte2016TeenTT, blakemore2018inventing}. Moreover, digital platforms, social media and gaming communities accelerate the spread of lexical experimentation and innovations, which go viral in youth networks at an unprecedented rate \cite{Eckert2012}. Therefore, young people, especially those under 25 \cite{grieve2018}, are perfect candidates for linguistic creativity testing. 

Regarding the general linguistic creativity of Large Language Models (LLMs), there is little or no scientific research. Most research papers focus only on one dimension of linguistic creativity, such as metaphors \cite{Chakrabarty2023ISA}, similes \cite{Chakrabarty2022FLUTEFL}, word formation \cite{Körtvélyessy_Štekauer_Kačmár_2022}, novel compounds \cite{coil-shwartz-2023-chocolate}, neologisms \cite{Lencione2022}, or morphological productivity \cite{anh-etal-2024-morphology, ismayilzada2025evaluatingmorphologicalcompositionalgeneralization}. However, a general comprehensive test for LLMs language creativity is missing from the literature. To bridge this gap, our paper aims to introduce such a test and to address the following research question: To what extent can LLMs deal with and adapt to creative language like slang, neologisms, or expressions that break morpho-syntactic and semantic norms, a skill so important for the ease of expressive communication in humans (generation and understanding)? If they can do it on a par with humans or even better, then there are major implications for industry, copyright infringement, safety, etc. The answer to this question RQ highly impacts various NLP applications like content generation, chatbots, automated moderation, linguistic engineering, etc.

Starting from these considerations about human and LLM linguistic creativity, our work aims to assess their overall abilities to innovate language based on a specifically designed multiple-task test. 

To this end, we structured the paper as follows. We begin by presenting some related works on LLMs creativity testing. Then, we introduce the test, describing the structure, the data set, and the evaluation methodology, as well as the setup for statistical analysis and visualization. In the subsequent sections, we present the results in the form of the scores obtained by humans and LLMs and their interpretation, also showcasing the uniqueness of the answers to the language creativity test. Finally, we draw some empirical remarks on the response database and formulate the general conclusions, including also the limitations and ethical compliance of our work.

\section{Related work}

A comprehensive survey on computational creativity including LLMs' various creativity types (problem solving, image, sound, ideational creativity, etc.) is given in \cite{ismayilzada2024creativityaiprogresseschallenges}. The LLM's creative writing abilities were tested in \cite{Chakrabarty2023ArtOA}. 

An integrated ideational creativity test was developed in \cite{DINU20242902}, in which the LLMs obtained slightly better scores than native English speakers on most of the test tasks. A more extensive testing of LLMs' ideational creativity \cite{Hubert2024TheCS} on a bigger dataset of native-speakers confirmed the results in \cite{DINU20242902}.

The overall language creativity of LLMs is less studied than other aspects of their creativity, like the ideational creativity mentioned above, in the sense that studies in this area focus on particular aspects. \cite{Chakrabarty2023ISA} pointed to LLMs' ability to generate metaphorical text descriptions of images, serving as input for diffusion-based text-to-image models. \cite{Chakrabarty2022FLUTEFL} used LLMs to help create a 9000 instance figurative language database focused on similes with explanations. \cite{coil-shwartz-2023-chocolate} employed a LLM to interpret and conceptualize novel noun compounds by generating explicit paraphrases. The model performed nearly perfectly on standard cases and outperformed humans on novel compounds, showing good generalization abilities. They also analyzed whether it reasons or merely echoes its training data. Moreover, \cite{Lencione2022} used LLMs to generate creative neologisms in a transfer learning setup. \cite{anh-etal-2024-morphology} tested LLMs with a multilingual Wug Test, assessing their ability to generalize morphological rules in six languages. The outputs were evaluated by native speakers, revealing that success depended on morphological complexity rather than training data size.

In a recent article, we developed a first language creativity test for humans and LLMs, as a pilot study \cite{dinu-florescu-2025-testing}. The current work improves the methodology (the first test proved to be more time consuming than initially expected, it was too complex, and there were some overlaps between language and ideational type of tasks), the criteria used (the \textit{Fluency} criterion was a major drawback for humans who were suspected that did not fully engage in generating the maximum number of ten answers, so it was eliminated) and the number of participants, both humans and LLMs.

\section{Test design}

The test we propose in this paper was designed to evaluate how creatively and originally humans and LLMs can manipulate language. The overall aim is to measure their ability to invent new words and use figurative language in unique ways, providing insights into their linguistic and imaginative capabilities. For human respondents, the test is meant to make them think outside the box and break away from conventional language use, encouraging them to play with language, experiment with word formation (i.e., derivation and compounding), which is seen as a creative act of a language user \cite{KörtvélyessyŠtekauerKačmár+2021+1017+1055}, explore metaphors, humor, and connotations (generating pejorative, neutral, or appreciative expressions for a given concept). For LLMs, the test serves as a benchmark to evaluate their ability to come up with contextually appropriate new words and figures of speech that go beyond the already learned patterns from the training data  \cite{boden2004creative,Boden_2009,Colton2012}. 
 
The test is structured into two parts, \textit{Word formation} and \textit{Metaphorical language}, each part containing four tasks. The tasks contain two items, for which the participants were required to provide three original answers, made up of up to five words. 

Part 1 (Word formation) focused on the creation of new words through word formation processes (derivation, compounding, and blending). The first task required the participants to combine two words to form a new one that describes a specific concept (e.g. "a person completely relying on chatbots"). The second task asked the respondents to complete a given word (like \textit{ice}) with another word to form a new compound (such as  \textit{icetomb}). Both tasks encouraged the use of synonyms and the creative blending of words to form meaningful and original compounds. The third task was to continue a certain series of derived words (like \textit{slangish}) using the same word formation process. This task targeted the respondents' ability to recognize and replicate linguistic patterns in creating new words. The last task of Part I provided a set of words related to a specific semantic field (like "dog, doggy, puppy, canine, bark, paw, woof") and asked participants to invent new words that would fit within the same semantic field (such as \textit{adogable}). This stimulated and tested the participants’ linguistic flexibility.

Part II (Metaphorical language) focused on figurative language, metaphors, and creative expressions, and challenged respondents to think metaphorically and come up with original, humorous, or thought-provoking phrases. The first task asked them to describe familiar ideas in new ways, giving them original/metaphorical alternative names (like \textit{fur missiles} for "angry small dogs"). The second task required the respondents to complete blank spaces so as to create an original metaphorical meaning (e.g. \textit{some...white hair strands...later}). In the third task, participants had to continue the given words with unusual or funny opposites (e.g. \textit{cozy...sitting on nails}). Finally, the respondents’ ability to manipulate tone and connotation while maintaining creativity and originality was tested in the last task of the test in which they were asked to give one original harsher (pejorative) expression, one equivalent expression, and one milder (appreciative) expression for a word (such as \textit{social storyteller} or \textit{muddy tongue} for "gossiper"). 

The test was drawn up in English and made available in Google forms. It was administered in controlled settings, in the classroom, under the supervision of the tutor, so as to instantly clarify any misunderstandings and avoid the use of digital aid. The answers were anonymously collected and processed in an aggregated form, in compliance with the ethical standards of scientific research. 

The test was administered to 24 volunteer students in Humanities (English major). They were all non-native speakers of English, with English language proficiency level of B2 and above. This English level ensures a solid grasp of English grammar and vocabulary, allowing them to engage in complex and nuanced language tasks. Aged between 19 and 25, the respondents, 8 males and 16 females, represent an optimal sample for the language creativity test, since, as stated in the introduction, young people tend to generate new words and meanings with greater ease \cite{Labov2001, Tagliamonte2016TeenTT, grieve2018}, a phenomenon that is further amplified by digital communication and social networking. 

We also administered the test to a set of 24 LLMs (20 with default settings and 4 with maximum \textit{top p} and \textit{temperature}, for the cases these features were available). All LLMs included in our study  are open-source.  Some  of the LLMs were used from their direct website: Claude (free version)\footnote{https://claude.ai/new}, Copilot\footnote{https://www.bing.com/chat?form=NTPCHB}, ChatGPT (free version)\footnote{https://chatgpt.com/}, Gemini (free version)\footnote{https://gemini.google.com/app}, Jais-30B\footnote{https://auth.arabic-gpt.ai/}, You.com (Smart mode)\footnote{https://you.com/?chatMode=default}, and  Falcon (180b)\footnote{https://huggingface.co/spaces/tiiuae/falcon-180b-demo}. Some were accessed via Huggingchat platform\footnote{https://huggingface.co/chat/models/}: DeepseekR1, Phi3.5, Cohere, Llama, and Qwen. Others were accesed from DeepInfra platform\footnote{https://deepinfra.com/}: Mistral nemotron, and WizzardLLM. Finally, others were used via Hugging Spaces platform\footnote{https://huggingface.co/spaces}: o3-mini, MiniMax, SOLAR, Hunyuan, databricks, and GrypheMythomax.

\section{Dataset}

Overall, our language creativity database, comprising answers from 24 humans and 24 LLMs, contains 2304 answers, (8 tasks x 2 items x 3 answers x 24 individuals = 1152 answers for each group, humans and LLMs), totaling 3904 words. The data is balanced in terms of the number of answers, but slightly unbalanced in terms of the number of words, with 2010 words for the LLM group and 1894 words for the human group. 
  
We eliminated formatting errors, namely any additional punctuation arbitrarily introduced by individuals, like quotation marks, numbers, or em-dash lines.

\section{Methodology and setup}

As postulated by \cite{guilford1950}, creativity assessment takes into account four criteria: \textit{Originality}, \textit{Flexibility}, \textit{Elaboration} and \textit{Fluency}. \textit{Originality} represents the uniqueness of the given answers. \textit{Flexibility} indicates how semantically distant the given answers are. \textit{Elaboration} measures the degree of detail of the responses. Finally, \textit{Fluency} keeps track of the number of ideas one comes up with \cite{alma991656963502466}.

To score the answers to the test, we employed a tool specifically designed to score creativity tasks, based on semantic distances and integration of GPT4, called Open Creativity Scoring with Artificial Intelligence (OCSAI version 1.6), which was shown to have a high correlation with human judgment \cite{ORGANISCIAK2023101356}. We used the full prompt option and the metaphor task type for scoring the\textit{ Originality} of the answers. We uploaded an Excel sheet containing the task prompt from the test and the corresponding answer. We normalized the scores provided by OCSAI from 1 to 5, to fit the 0-1 interval, 0 being the least original. The \textit{Fluency} criterion is irrelevant for this test, since we requested a fixed mandatory number of answers, namely three. This choice is motivated by the fact that LLMs would invariably respond with exactly the number of answers required by the test, however large that number would be, while humans do not parallel such unlimited resources and their number of answers would greatly depend on their motivation rather than their actual capabilities, resulting in a biased overall score in favor of the LLMs. \textit{ Elaboration} was assessed by OCSAI via word count limit.  Since OCSAI does not have the option for scoring \textit{Flexibility}, which is an important criterion of creativity assessment, we generated all pairs of answers given by an individual to a particular task item and we subsequently scored them with OCSAI. The final \textit{Flexibility} score was obtained as the mean of the scores for all these pairs of answers.

For data and statistical analysis, and for visualization, we used ChatGPT (4o)\footnote{https://chatgpt.com/} to assist with coding, and Python with the following libraries: nltk\footnote{https://www.nltk.org/}, scipy\footnote{https://scipy.org/}, numpy\footnote{https://numpy.org/}, matplotlib\footnote{https://matplotlib.org/}, seaborn\footnote{https://seaborn.pydata.org/}, and pandas\footnote{https://pandas.pydata.org/}. 
 
\section{Results}

\begin{figure}
    \centering
    \begin{minipage}{0.49\linewidth}
        \centering
        \includegraphics[width=\linewidth]{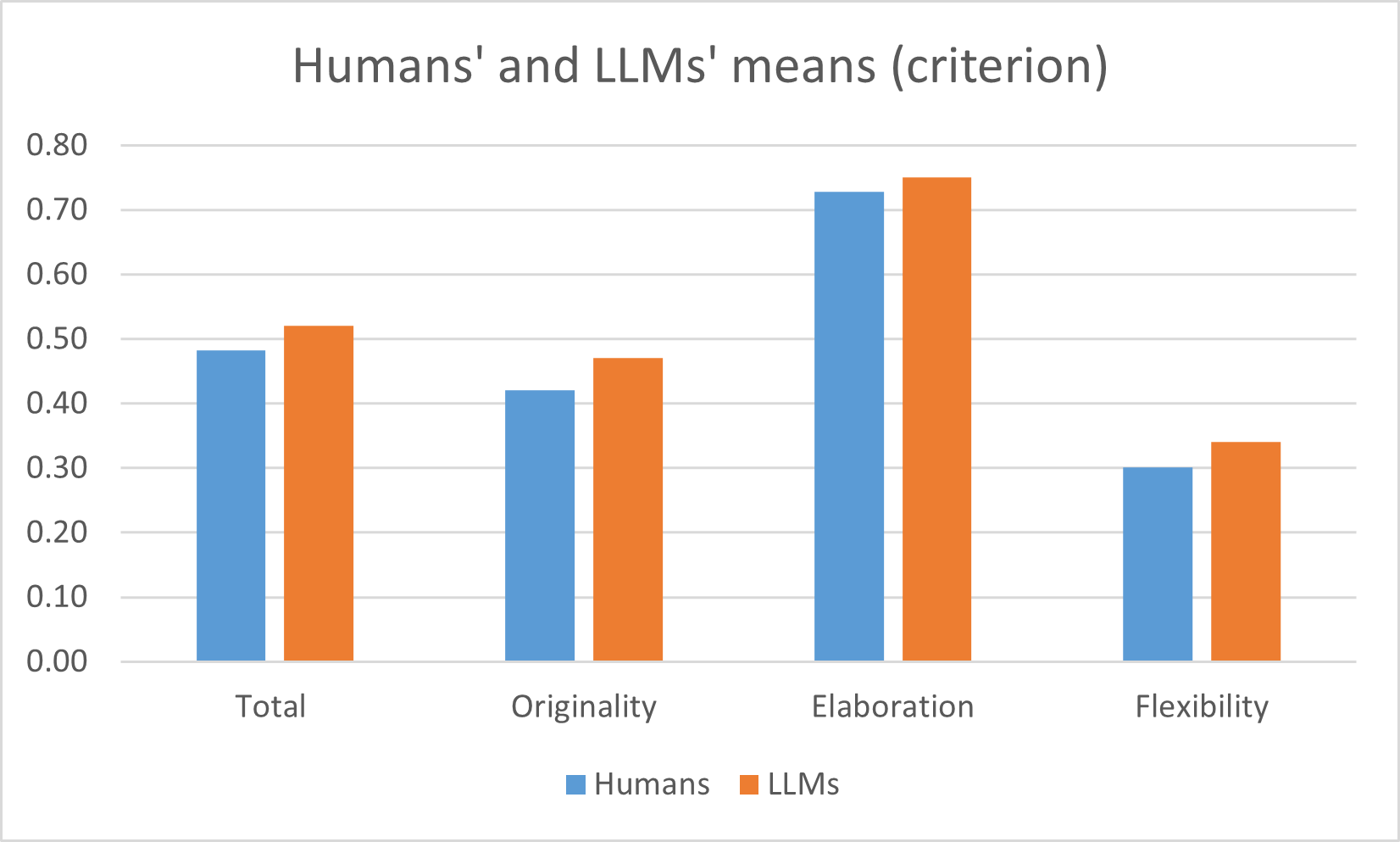}
        \caption{Humans’ versus LLMs’ mean scores per criterion}
        \label{fig:meanscriterion}
    \end{minipage}
    \hfill
    \begin{minipage}{0.49\linewidth}
        \centering
        \includegraphics[width=\linewidth]{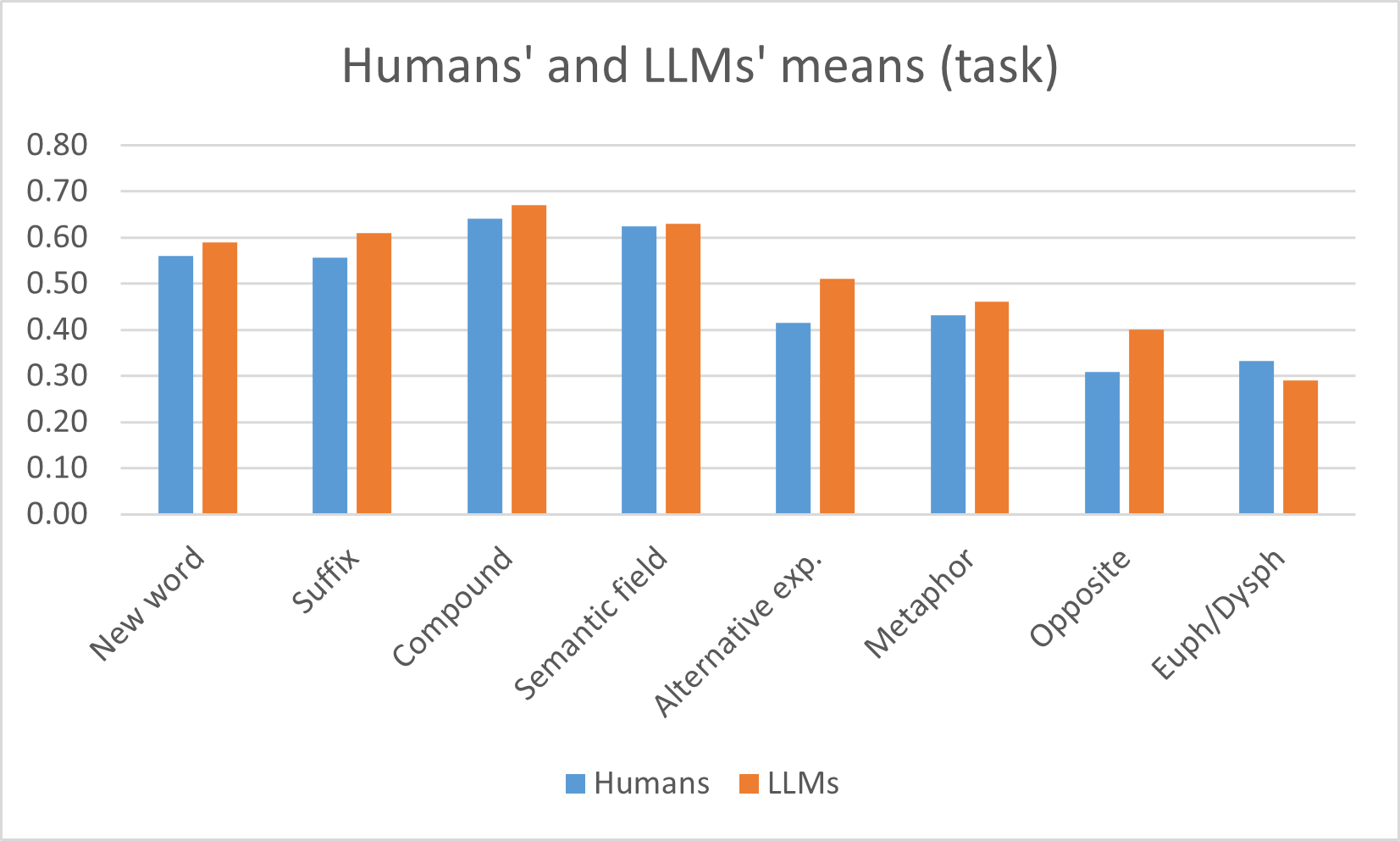}
        \caption{Humans’ versus LLMs’ mean scores per task}
        \label{fig:meanstas}
    \end{minipage}
\end{figure}

We computed the average scores for each individual (human and LLM) over all their answers for the three criteria, the eight tasks, and the total. The results, ordered in decreasing order of their total scores, are shown in Table \ref{tab:human_llm_scores} for both humans and LLMs. 
We then calculated the overall mean for humans and LLMs, per criterion, and per task, represented in figure \ref{fig:meanscriterion}, and \ref{fig:meanstas}, respectively. The total mean for the LLMs as a group was 0.52, while the humans obtained a mean of 0.48. The LLMs outperformed humans not only for all three criteria, but almost for all tasks, with the exception of \textit{Semantic Field} task, where they were on a par, and of Euphemism/Dysphemism, where humans slightly outperformed LLMs with 0.33 to 0.29.

\begin{figure}
    \centering
    \begin{minipage}{0.49\linewidth}
        \centering
        \includegraphics[width=\linewidth]{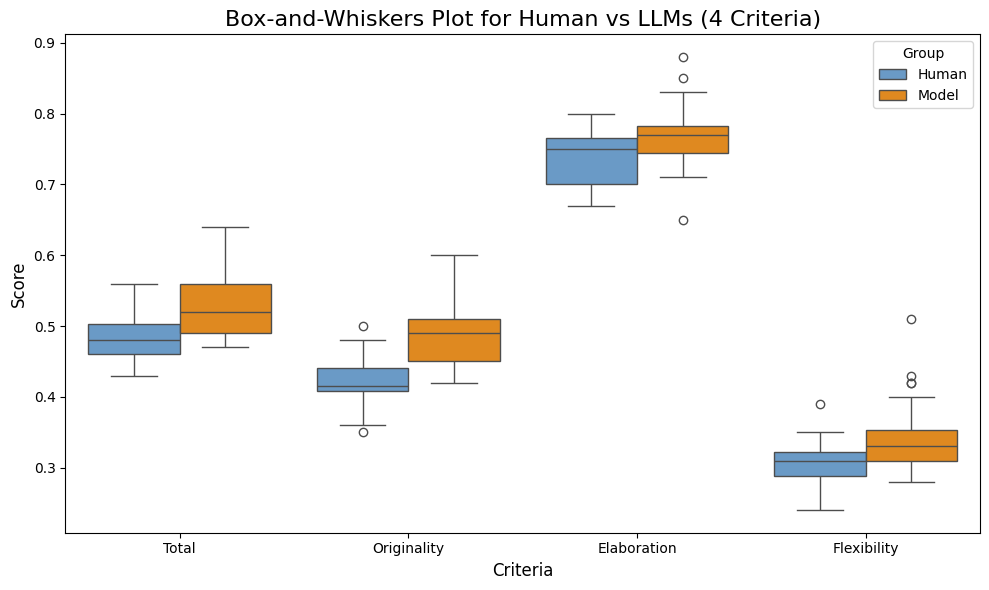}
        \caption{Statistics for Humans’ versus LLMs’ mean scores per criterion}
        \label{fig:whiskers}
    \end{minipage}
    \hfill
    \begin{minipage}{0.49\linewidth}
        \centering
        \includegraphics[width=\linewidth]{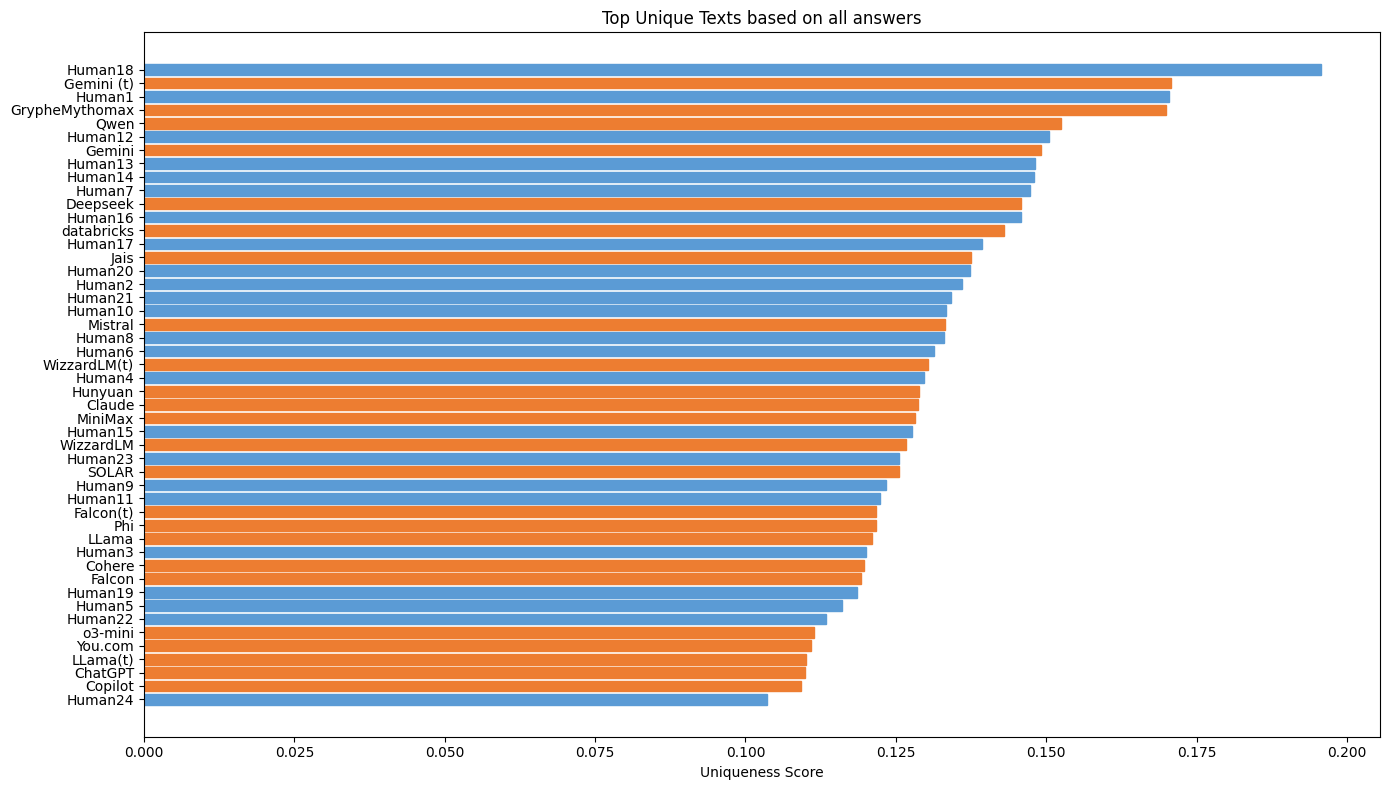}
        \caption{Top uniqueness of the dataset}
        \label{fig:uniqueness}
    \end{minipage}
\end{figure}

The statistics for the overall creativity and the three criteria scores are illustrated in figure \ref{fig:whiskers}. The total mean scores are higher for LLMs, while the variance is higher among LLMs than among humans. For the \textit{Originality} criterion, the LLMs clearly outperformed humans. The LLM mean scores are also more spread out compared to human scores. Moreover, there are some outliers in the human group, one who performed worse and one better. For \textit{Elaboration}, the median is a little higher for LLMs than for humans, but the variance is higher this time in humans than in the LLM group. Also, there are three LLM outliers, two of them overperforming and one underperforming. As for the \textit{Flexibility} criterion, the LLMs performed better than humans again, whereas the variances of the two groups are similar. Here, there is a human overperformer and three LLMs. The figure visually confirms that LLMs generally outperformed humans in \textit{Total} mean scores and in all three criteria. However, the overlap in distributions suggests that some humans achieved scores comparable to LLMs, meaning performance isn't entirely one-sided.

To check the statistical significance of the differences in mean scores for the two groups, per criterion and in total, we performed a t-test. The results are as follows: 
\begin{itemize}
    \item for the \textit{Total} score, the t-statistic value is -3.03 (the difference between the group means is approximately 3 standard deviations from 0 in favor of the LLMs) and p-value is 0.0044 (the differences between the group means is statistically significant); 
    \item for the \textit{Originality} score, the t-statistic value is -4.21 (the difference between the group means is approximately 4 standard deviations from 0 in favor of the LLMs) and p-value is 0.00014 (the differences between the group means is very statistically significant); 
    \item for the \textit{Elaboration} score, the t-statistic value is -1.15 (the difference between the group means is approximately one standard deviation from 0 in favor of the LLMs) and p-value is 0.258 (the differences between the group means is not statistically significant); 
    \item for the \textit{Flexibility} score, the t-statistic value is -3.06 (the difference between the group means is approximately 3 standard deviations from 0 in favor of the LLMs) and p-value is 0.0040 (the differences between the group means is statistically significant).
\end{itemize}

In addition, to test the statistical significance of the differences between humans' and LLMs' scores using all available data (48 individuals x 3 criteria x 8 tasks = 1152 scores), and not only the mean scores per individual, we used an Ordinary Least Squares (OLS) regression model. We analyzed creativity scores as a function of the group (Humans vs. LLMs). The results showed a small but statistically significant effect of the group (p = 0.016), with LLMs scoring on average 0.037 points higher than Humans.

The DeepSeek model leads at a comfortable distance from the rest of the models. 

The effect of maximizing the \textit{temperature} and \textit{top p} for the four models where they were available was inconclusive: two models showed some improvement of the total creativity score (Gemini and Falcon), while the other two underperformed (WizzardLM and Llama).

\begin{table}
  \caption{Scores based on criterion ordered by Total (Humans and LLMs)}
  \centering
  \begin{subtable}{0.49\linewidth}
    \centering
    \caption{Humans}
    \begin{tabular}{lcccc}
      \toprule
      \textbf{Human} & \textbf{Total} & \textbf{Orig.} & \textbf{Elab.} & \textbf{Flex.} \\
      \midrule
      Human14 & 0.56 & 0.50 & 0.79 & 0.39 \\
      Human12 & 0.53 & 0.47 & 0.80 & 0.32 \\
      Human5  & 0.52 & 0.45 & 0.80 & 0.33 \\
      Human16 & 0.52 & 0.48 & 0.73 & 0.35 \\
      Human6  & 0.51 & 0.45 & 0.76 & 0.32 \\
      Human3  & 0.51 & 0.43 & 0.75 & 0.34 \\
      Human24 & 0.50 & 0.44 & 0.76 & 0.29 \\
      Human15 & 0.50 & 0.42 & 0.78 & 0.29 \\
      Human17 & 0.49 & 0.43 & 0.75 & 0.31 \\
      Human13 & 0.49 & 0.41 & 0.70 & 0.35 \\
      Human10 & 0.48 & 0.44 & 0.70 & 0.31 \\
      Human18 & 0.48 & 0.41 & 0.76 & 0.26 \\
      Human21 & 0.48 & 0.42 & 0.74 & 0.28 \\
      Human4  & 0.48 & 0.43 & 0.67 & 0.34 \\
      Human22 & 0.47 & 0.39 & 0.78 & 0.24 \\
      Human11 & 0.47 & 0.41 & 0.71 & 0.28 \\
      Human9  & 0.46 & 0.41 & 0.68 & 0.29 \\
      Human7  & 0.46 & 0.41 & 0.67 & 0.31 \\
      Human2  & 0.46 & 0.40 & 0.68 & 0.29 \\
      Human8  & 0.45 & 0.38 & 0.70 & 0.28 \\
      Human23 & 0.45 & 0.40 & 0.69 & 0.27 \\
      Human1  & 0.45 & 0.36 & 0.74 & 0.25 \\
      Human19 & 0.44 & 0.40 & 0.65 & 0.29 \\
      Human20 & 0.43 & 0.35 & 0.70 & 0.26 \\
      \bottomrule
    \end{tabular}
  \end{subtable}%
  \hfill
  \begin{subtable}{0.49\linewidth}
    \centering
    \caption{LLMs}
    \begin{tabular}{lcccc}
      \toprule
      \textbf{Model} & \textbf{Total} & \textbf{Orig.} & \textbf{Elab.} & \textbf{Flex.} \\
      \midrule
      Deepseek & 0.64 & 0.60 & 0.82 & 0.51 \\
      WizzardLM & 0.59 & 0.54 & 0.81 & 0.43 \\
      Llama & 0.58 & 0.53 & 0.88 & 0.33 \\
      WizzardLM(t) & 0.57 & 0.52 & 0.78 & 0.40 \\
      Cohere & 0.56 & 0.49 & 0.79 & 0.42 \\
      You.com & 0.56 & 0.50 & 0.83 & 0.36 \\
      MINIMax & 0.56 & 0.52 & 0.75 & 0.42 \\
      o3-mini & 0.55 & 0.49 & 0.85 & 0.33 \\
      Claude & 0.54 & 0.50 & 0.78 & 0.34 \\
      Gemini(t) & 0.53 & 0.50 & 0.77 & 0.32 \\
      Falcon(t) & 0.53 & 0.51 & 0.76 & 0.31 \\
      Llama(t) & 0.52 & 0.49 & 0.77 & 0.30 \\
      ChatGPT & 0.52 & 0.47 & 0.73 & 0.35 \\
      SOLAR & 0.51 & 0.45 & 0.72 & 0.37 \\
      Mistral & 0.51 & 0.51 & 0.71 & 0.32 \\
      Copilot & 0.51 & 0.44 & 0.75 & 0.34 \\
      Phi & 0.49 & 0.45 & 0.72 & 0.31 \\
      Hunyuan & 0.49 & 0.43 & 0.70 & 0.34 \\
      Falcon & 0.48 & 0.42 & 0.73 & 0.31 \\
      Gemini & 0.48 & 0.44 & 0.69 & 0.31 \\
      databricksdbrx & 0.46 & 0.41 & 0.70 & 0.27 \\
      Jais & 0.45 & 0.40 & 0.65 & 0.29 \\
      GrypheMythomax & 0.43 & 0.37 & 0.65 & 0.29 \\
      Qwen & 0.43 & 0.42 & 0.60 & 0.28 \\
      \bottomrule
    \end{tabular}
  \end{subtable}
  \label{tab:human_llm_scores}
\end{table}

\section{Uniqueness}

 Since one criterion for evaluating creativity in psychology refers to how original an individual's responses are in comparison with those of others \cite{carayannis2013encyclopedia}, we computed the uniqueness score of all individuals. This meant that we assessed the differences between each individual's answers and the answers of all the other individuals. Following the same methodology as in \cite{DinuF24}, we grouped the answers to the test of both humans and LLMs in separate files, each containing all the answers of a particular individual. We thus obtained 48 answer files, 24 for humans and 24 for LLMs. 
 
 We removed the stop words with NLTK library, since we compared linguistic originality, and not individual style, and stop words are good discriminants in authorship attribution \cite{dinu-uban-2023-computational}. We then generated embeddings for each file and we computed their pairwise semantic similarity, using spaCY library. The uniqueness scores were obtained as the inverse of the average semantic similarity scores between an individual and all the others. 
 
 In figure \ref{fig:uniqueness}, we list the uniqueness scores obtained, in decreasing order. There are no obvious patterns that differentiate LLMs from humans, except for a slightly more predominant occurrence of humans in the first part of the ranking (15 humans and 9 LLMs), and, symmetrically, of LLMs in the second part (15 LLMs and 9 humans). Furthermore, the most unique and the least unique individuals were both humans, with uniqueness scores considerably different from the next/previous scores, indicating that more humans than machines tend to be quirky w.r.t. their linguistic creativity.

Since the most unique individual was by far Human 18, we manually inspected his answers. We observed that almost all of them were indeed not present in any other set of individual answers. Even if he was not scored by OCSAI with high total and originality scores (he is in the middle of the human ranking based on the total creativity score), Human 18 responded with unique answers w.r.t. all the other answers in the database. For instance, his answers included \textit{LLMoron} for a "a person completely relying on chatbots", \textit{contentmess} for "a mixt feeling of satisfaction and guilt", \textit{barkative} and \textit{hogpot} for the task of creating a new word in the semantic field of \textit{dog} and \textit{pig}, respectively, \textit{some... head scratches... later} and \textit{a sea of... intrusive thoughts} for filling in the blanks with original metaphors.

\section{Empirical remarks}

A ‘naked eye’ comparison of the humans’ and LLMs’ responses in the given linguistic creativity tasks reveals some differences in cognitive strategies, rule adherence, and pragmatic intent. These differences are more evident across the tasks requiring word formation, metaphorical naming, and morphological play. 

For example, in Part I, task 1, the respondents were asked to name “a person completely relying on chatbots”. On the one hand, LLMs demonstrated a more rule-based creativity,  resorting to a variety of standard word formation processes when creating nonce words, which resulted from the combination of the given words or of their close synonyms (e.g. compounding: \textit{chat-dependent, botdependent, botleaner, chatreliant}; truncation: \textit{chatabotnik, persobot}; derivation: \textit{chatbotter, botixian}; blending: \textit{chatrivore, chatrogate}). In that regard, the LLMs seemed to prioritize F-creativity.  On the other hand, humans created compounds or blends by merging more semantically distant words, demonstrating more freedom in the selection of items from the same lexical field (e.g. z\textit{ombonoid, robotinator, stupiding, cyberdude, cyberstudents, chatversed}), some of which having a metaphorical construal (e.g. \textit{braindeadman, socialblind}), showing a tendency towards E-creativity. Another example comes from Part I, task 2. Here respondents had to create original, novel derived words with the suffixes -\textit{ish} and -\textit{able}. LLMs attached the suffix \textit{-ish} mostly to adjectives or nouns, ‘playing safe’ in the game of derivation, whereas the humans offered ‘rule-breaking’ examples by attaching it to non-standard bases, like verbs (e.g. \textit{scrollish, \textit{hatish, loveish}}), prioritizing expressivity over morphological rules, and often violating expected patterns for playful or ironic effect (e.g. creating words like \textit{wifeable, schoolable} with the suffix -\textit{able}). Again the humans were more prone to E-creativity than the LLMs.

Moreover, humans demonstrated in some cases a rather intensional deviation, especially in the responses in which they seemed more inclined to create playful combinations with pragmatic functions (expressivity, humor, word play, metaphorical use, attention seekers, even offensive language). For instance, task 1 from Part II targeted alternative, metaphorical names for “angry small dogs”, which human respondents called \textit{backing vocals from the back yard} or \textit{electrocuted hotdogs}, and for "the I wanna speak to the manager kind of people", for which they used offensive words like \textit{dickheads}. LLMs, on the other hand, created  statistically plausible but socially inert outputs like \textit{miniature furry clouds, pocket-size volcanoes, furious pint-sized furries} for “angry small dogs”, with words or phrases repeating in many answers (\textit{fury, tiny, furry, miniature, pocket-sized, tempest, tornado}) and numerous alliterations (\textit{tiny temper tantrum terrors, tiny tempest terriers, fiery furball fighters, snarling socks}), most likely because alliterations seen in the training set functioned in context as a means of increasing the emotional intensity and of reproducing the audible effect of the given phrase. 

Last but not least, LLMs outperformed humans in rule-based morphological play, generating more numerous and structurally varied nonce formations when asked to create new words in a particular semantic field, such as "dog" or "pig", producing combinations like \textit{pawtato, pupparazzi, barktastic, snifftacular, pawzigraphy, canifrolic}, or \textit{oinkectomy, oinksnuff, oinktastic, oinktharsis, oinkery}. 

Overall, while humans demonstrated greater flexibility and intentionality in their linguistic creativity, employing \textit{flair} \cite{Franceschelli2024} through more contextually-grounded and pragmatically-rich combinations, their output proved less systematically productive than the LLMs'. This paradox aligns with Chomsky's \cite{chomsky2003reasons} observation that language operates within fixed laws yet enables infinite variation.

\section{Conclusions}

We introduced a linguistic creativity test designed to assess and compare LLM versus human abilities to use language creatively through word formation processes and metaphorical construal. 

We tested 24 humans and 24 LLMs, thus obtaining a balanced dataset of 2304 creativity test answers, which we automatically evaluated. The results showed that LLMs outperformed humans in all criteria and in most of the tasks. Moreover, the human group  proved slightly less homogeneous w.r.t. uniqueness scores than the LLM group.

Finally, manual observation revealed that humans seem more inclined towards E-creativity, often bending linguistic rules, while LLMs favor F-creativity, covering a great variety of rule-based instances.

Overall, the results of this study suggest that LLMs possess a human-like capacity to generate contextually appropriate original words and expressions. This has important implications for knowledge engineering: systems that can creatively manipulate language at a morphological and metaphorical level are better equipped to handle informal, evolving, or domain-specific vocabularies, including slang, neologisms, or nonce words and expressions. Such capabilities are essential for robust knowledge extraction, flexible ontology development, and improved semantic interoperability in dynamic real-world applications. The ability of LLMs to not only interpret, but also generate previously unseen terms indicates their potential to support more adaptive and cognitively plausible knowledge systems.

\section{Limitations and future work}

One limitation of this study is the lack of native respondents, and thus the question of how different the results would have been if native speakers had been included remains to be answered.
In a study on ideational creativity of native English speakers, \cite{Hubert2024TheCS} reached similar results as \cite{DINU20242902}, who used non-native English speakers. Both works showed that LLMs scored higher than humans, when automatically evaluated with the same tool we used in this article, OCSAI. However, in this study, we assessed linguistic creativity, and not ideational creativity, so, in this case, significant differences in the scores obtained by humans and LLMs could occur.  

In our future work, we will enhance the data set by including answers collected from native speakers. Moreover, we will add a more in-depth analysis of the dataset, such as sentiment analysis, clustering, human-machine classification of language creativity answers, etc.

\section{Ethical statement}
This study adheres to the ethical guidelines for academic research established by the University of Craiova. The test design, methodology, and data collection procedures were reviewed and approved by the University of Craiova Ethics Commission (Decision No. 481/21.02.2025). All participants provided informed consent prior to their involvement, with explicit clarification regarding the purpose of the study and their role in it, the voluntary nature of participation and their right to withdraw at any stage without consequence, as well as the anonymity and confidentiality of their responses, ensuring that no personally identifiable information is disclosed in publications or reports. The research complies with the principles of integrity, transparency, and respect for participants' autonomy, aligning with international ethical standards for human-subject research (e.g., GDPR for EU-based participants).

\section*{Acknowledgments}

This work was supported by a grant of the Ministry of Research, Innovation and Digitization, CNCS-UEFISCDI, project SIROLA, number PN-IV-P1-PCE-2023-1701, within PNCDI IV.
We also acknowledge the support of our students who participated in the test and we thank them for their valuable and creative input and for their enthusiasm.

\bibliographystyle{unsrt}  
\bibliography{references}  %%% Remove comment to use the external .bib file (using bibtex).

% %%% Comment out this section when you \bibliography{references} is enabled.

% George Kour and Raid Saabne.
% \newblock Real-time segmentation of on-line handwritten arabic script.
% \newblock In {\em Frontiers in Handwriting Recognition (ICFHR), 2014 14th
%   International Conference on}, pages 417--422. IEEE, 2014.

% George Kour and Raid Saabne.
% \newblock Fast classification of handwritten on-line arabic characters.
% \newblock In {\em Soft Computing and Pattern Recognition (SoCPaR), 2014 6th
%   International Conference of}, pages 312--318. IEEE, 2014.

% \bibitem{hadash2018estimate}
% Guy Hadash, Einat Kermany, Boaz Carmeli, Ofer Lavi, George Kour, and Alon
% \newblock Estimate and replace: A novel approach to integrating deep neural
%   networks with existing applications.
% \newblock {\em arXiv preprint arXiv:1804.09028}, 2018.

%\end{thebibliography}

\end{document}